\documentclass[runningheads]{llncs}
\usepackage{amsmath,amssymb,amsfonts}

\usepackage{marvosym}
\usepackage{cite}
\usepackage{algorithmic}
\usepackage{url}    
\usepackage{graphicx}
\usepackage{textcomp}
\usepackage{xcolor}
\usepackage{graphicx}
\usepackage{amsmath}
\usepackage{float}
\usepackage{setspace}

\AtBeginDocument{%
  \setlength{\abovedisplayskip}{3mm}
  \setlength{\belowdisplayskip}{3mm}
}

\usepackage[pagebackref,breaklinks,colorlinks]{hyperref}
\hypersetup{
    colorlinks=true,     
    linkcolor=blue,      
    citecolor=blue,      
    urlcolor=blue,       
    linkbordercolor=white, 
    citebordercolor=white,
    pdfborder={0 0 0}    
}
\usepackage{subfigure}  
\usepackage{arydshln}  
\usepackage{rotating}
\usepackage{array}
\usepackage{tikz}
\usepackage{algorithm}
\usepackage{colortbl}
\usepackage{adjustbox}
\usepackage{calc}
\usepackage{makecell}
\usepackage{multirow}
\usepackage{booktabs}
\begin{document}
\title{FlexiGrad: Adaptive Gradient Modulation for Hierarchical Fine-Grained Classification}

\titlerunning{FlexiGrad for Hierarchical FGVC}

\author{Zilu Zhou\inst{1,2} \and
Dongliang Chang\inst{1,2,}\textsuperscript{\Letter} \and
Junhan Chen\inst{1,2} \and Zhanyu Ma\inst{1,2}}
\authorrunning{Z. Zhou et al.}

\institute{Beijing University of Posts and Telecommunications, Beijing 100876, China \and
Beijing Key Laboratory of Multimodal Data Intelligent Perception and Governance, Beijing 100876, China\\
\email{changdongliang@bupt.edu.cn}
}
\maketitle              
\begin{abstract}
Many fine-grained recognition tasks contain hierarchical labels such as order, family and species.  
Although this supervision should be beneficial, jointly optimising all levels often leads to unstable training because coarse and fine classifiers impose inconsistent gradients on the shared backbone.  
This hierarchical gradient conflict prevents the model from learning a coherent coarse-to-fine representation.
In this paper, we propose FlexiGrad, a simple and parameter-free method that regulates gradient interactions during backpropagation.  
FlexiGrad removes only the harmful conflicting component when tasks disagree and reinforces the shared direction when they partially agree through a smooth hierarchy-aware weighting function.  
This produces stable optimisation and preserves both global structure and fine-grained discriminative cues.
FlexiGrad integrates into existing architectures without modification while improves multi-granularity accuracy on CUB-200-2011, FGVC-Aircraft and Stanford Cars. The code will be available at \href{https://github.com/PRIS-CV/FlexiGrad}{PRIS-CV/FlexiGrad}.

\end{abstract}

\section{Introduction}

\begin{figure}[t]
\centering
\includegraphics[width=0.98\textwidth]{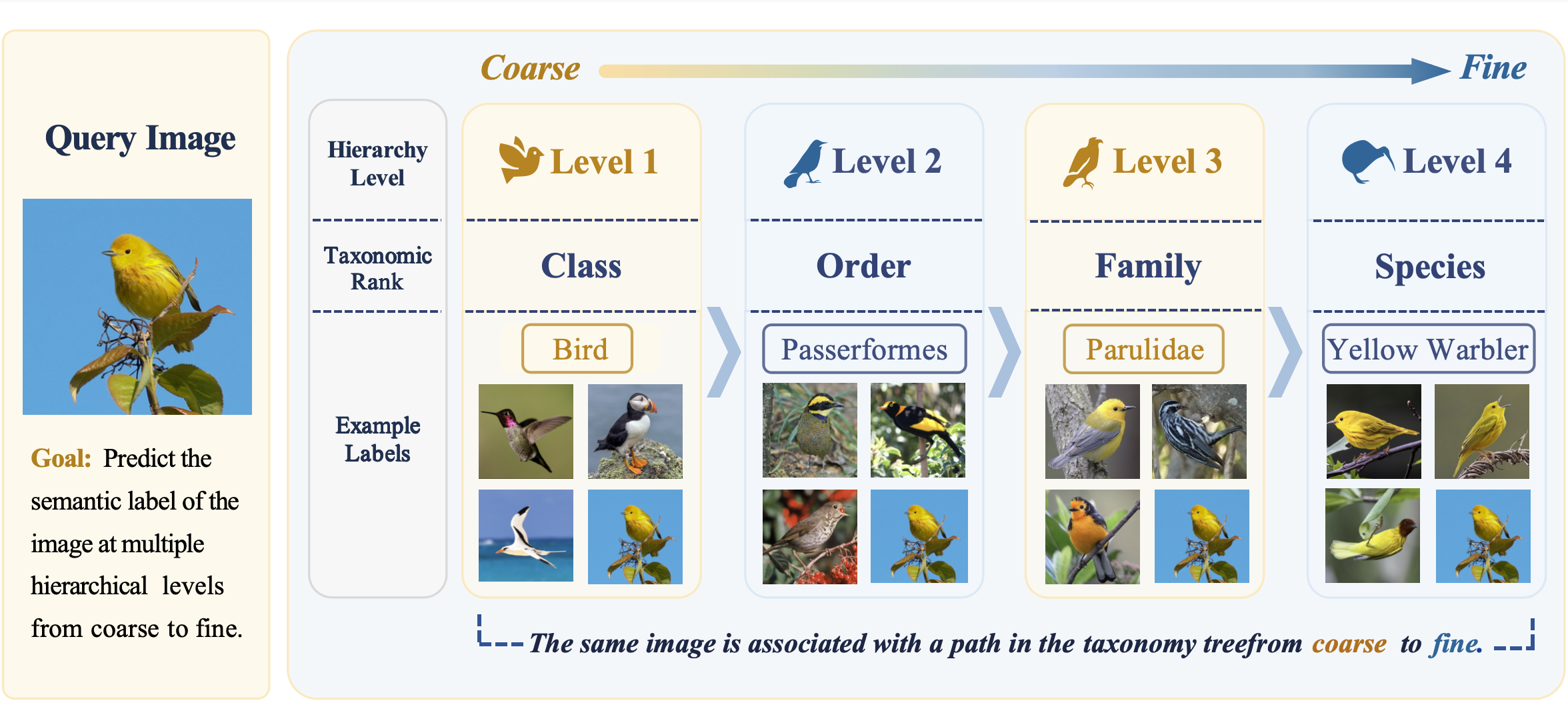}
\caption{Fine-Grained Visual Classification (FGVC) with Hierarchical Labels for Multi-Granularity Recognition.} \label{fig:1}
\vspace{-3mm}
\end{figure}

Hierarchical supervision is one of the most powerful forms of structure available in fine-grained recognition~\cite{du2021progressive}.  
A single image often carries multiple labels at different semantic levels such as order, family and species (as shown in Figure~\ref{fig:1})~\cite{wah2011caltech, maji2013fine, li2019dual}.  
Coarse labels define global appearance patterns and fine labels provide subtle discriminative cues.  
In theory, learning all levels jointly should make recognition easier, not harder.  
Yet in practice, models trained with multi-granularity supervision often converge more slowly, behave less stably and sometimes underperform models trained using only the finest labels.  
This paradox suggests that the challenge does not lie in the labels themselves, but in the way optimisation handles them.

\begin{figure}[t]
\centering
\includegraphics[width=0.93\textwidth]{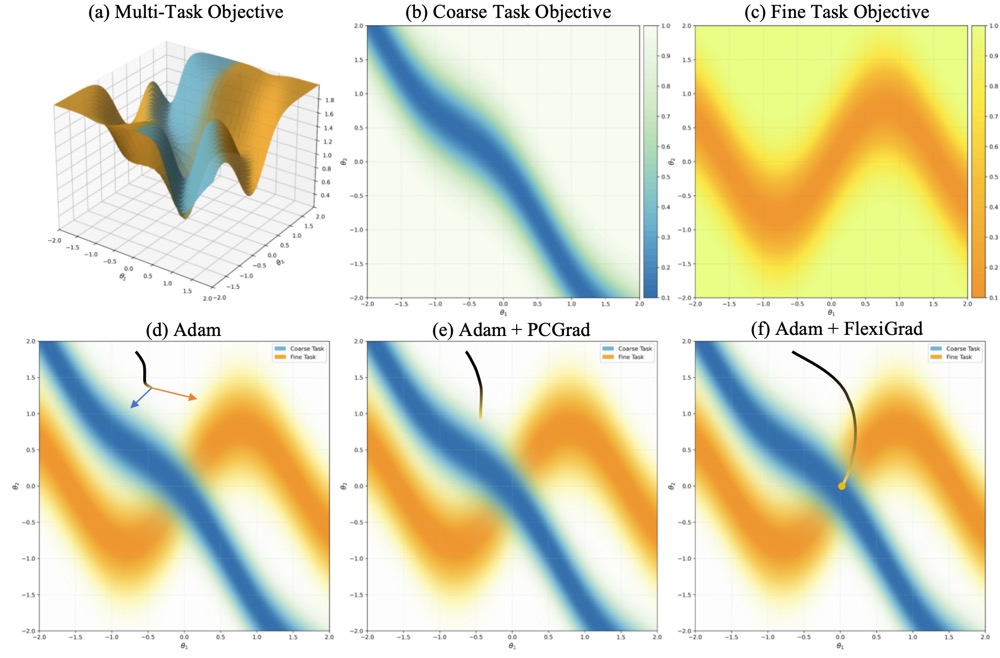}
\caption{(a) The 3D landscape of the combined multi-task (coarse \& fine Classification) objective. (b) \& (c) Contour plots of the individual Coarse Task and Fine Task objectives. (d) Trajectory using the standard Adam optimizer. The gradient vectors of the two tasks at the end of the trajectory are indicated by blue and orange arrows. (e) Trajectory using Adam with PCGrad. (f) Trajectory using Adam with FlexiGrad.} \label{fig:intro}
\vspace{-7mm}
\end{figure}

A multi-granularity setting naturally resembles multi-task learning because each granularity level defines a related but distinct objective.  
Although these objectives share the same backbone, they influence it in different and sometimes incompatible ways.  
Coarse classifiers encourage broad and shape-dominant features, while fine classifiers depend on highly localised and subtle attributes~\cite{chang2021your, kuang2026spatial, chang2020devil}.  
During backpropagation their gradients frequently point in inconsistent directions~\cite{yu2020gradient}, which creates hierarchical gradient conflict.  
This conflict distorts the shared parameter landscape and makes the joint objective significantly harder to optimise than any individual task.  
Even simple cases lead to twisted descent paths as shown in Figure~\ref{fig:intro}.  
The backbone receives competing signals, and the final representation fails to fully exploit the hierarchical structure that should have been beneficial.

Existing solutions attempt to solve this problem from two separate perspectives, but both fall short when applied to hierarchical fine-grained recognition.  
Gradient-modification methods such as PCGrad~\cite{yu2020gradient} and CAGrad~\cite{liu2021conflict} adjust gradients through geometric projections or balancing objectives.  
These techniques stabilise training for loosely related tasks, yet they treat all tasks symmetrically and therefore ignore the semantic roles encoded in a hierarchy.  
A coarse-level gradient that promotes global consistency should not be suppressed in the same way as a conflicting fine-level gradient that carries discriminative evidence.  
Uniform geometric treatment often eliminates signals that should guide the backbone, weakening the very information that hierarchical supervision is supposed to provide.

Architectural approaches take the opposite direction.  
Hierarchical FGVC models frequently restrict or block gradient flow across granularity levels to prevent negative interference~\cite{chang2021your}.  
Such designs avoid harmful interactions, but at the cost of removing positive transfer and limiting the backbone's ability to learn a coherent coarse-to-fine representation.  
Once gradients are isolated, the hierarchy becomes a set of disconnected prediction problems. The structural advantage of multi-level labels is largely lost.

These limitations point to a critical missing capability.  
A hierarchical system requires neither full symmetry nor full separation.  
It requires a mechanism that regulates how coarse and fine tasks influence one another, adjusts their interactions based on consistency and preserves the semantic priorities of the hierarchy.  
The optimisation process must become sensitive to these relationships instead of treating all disagreements as harmful or all agreements as equally valuable.

We introduce FlexiGrad, a simple and parameter-free gradient modulation mechanism for hierarchical fine-grained classification.  
FlexiGrad operates entirely during backpropagation and does not modify the network architecture.  
It computes the consistency between gradients from different granularity levels and adjusts their contribution adaptively.  
When two tasks conflict strongly, FlexiGrad removes only the harmful component and keeps the useful part.  
When two tasks share partial agreement, FlexiGrad reinforces their aligned direction using a smooth weighting function preserving their individual characteristics.  
This produces an optimisation trajectory that suppresses destructive interference while amplifying beneficial cooperation.  
Fine-level gradients naturally gain greater influence because they carry the most discriminative information, while coarse-level signals continue to stabilise global structure.

FlexiGrad integrates seamlessly with standard backbones and state-of-the-art FGVC frameworks.  
Across CUB-200-2011~\cite{wah2011caltech}, FGVC-Aircraft~\cite{maji2013fine} and Stanford Cars~\cite{krause2013car} it improves accuracy at all granularity levels and delivers the largest gains on difficult families and species where gradient conflict is most severe.  
Training becomes more stable, attention becomes more coherent and the backbone learns a representation that better reflects the true taxonomy of the data.

\section{Related Work}
\label{sec:background}

\textbf{Multi-Task Learning and Gradient Conflict} Multi-task learning (MTL) jointly trains multiple objectives with a shared backbone and task-specific heads~\cite{liu2025pareto}.  
Early work balanced task importance using static or adaptive loss weighting~\cite{kirchdorfer2026investigating}, while later approaches introduced architectural mechanisms such as task routing and soft parameter decoupling~\cite{matsubara2025multi, gao2026toward, gao2023self, qin2025multimodal, qin2026increfa}.

A central issue in MTL is gradient conflict.  
Gradients from different tasks may point in inconsistent directions, causing inter-task interference~\cite{abdelsamie2026deep}.  
To address this, several methods directly modify gradients during backpropagation.  
PCGrad~\cite{yu2020gradient} removes conflicting components through projection, and CAGrad~\cite{liu2021conflict} searches for a balanced descent direction.  
DRGrad~\cite{liu2025direct} further refine gradients using agreement-based rules.

These approaches improve stability but operate purely on gradient geometry, ignoring the semantic structure linking tasks.  
In hierarchical classification, coarse and fine objectives form meaningful dependencies rather than independent tasks.  
Treating their gradients without considering this relationship risks discarding useful information.  
Our method addresses this limitation by adjusting gradient interactions in a way that respects both their alignment and their hierarchical roles.

\textbf{Hierarchical FGVC Strategies} Hierarchical supervision has been incorporated into fine-grained visual classification through structural or architectural designs~\cite{chen2026seeing, wang2026detailverifybenchbenchmarkdensehallucination, wang2026cinetechbench, gao2025cross, chen2022cross, zheng2025hierarchical}.  
FGoN~\cite{chang2021your} blocks gradient flow between levels to avoid negative transfer, and other works apply class hierarchy priors~\cite{liu2025long} or hierarchical residual structures to guide feature learning.

These methods focus on feature design or architectural separation and often avoid explicit gradient interaction.  
Although blocking gradients prevents harmful interference, it also removes potentially beneficial cross-level signals and limits coarse-to-fine representation learning.

Our work follows a complementary direction.  
Instead of restricting or isolating gradients, we modulate them adaptively based on consistency.  
This allows the model to suppress harmful interactions while reinforcing helpful ones, essential in hierarchical FGVC where tasks are semantically linked.

\begin{figure}[t]
\includegraphics[width=\textwidth]{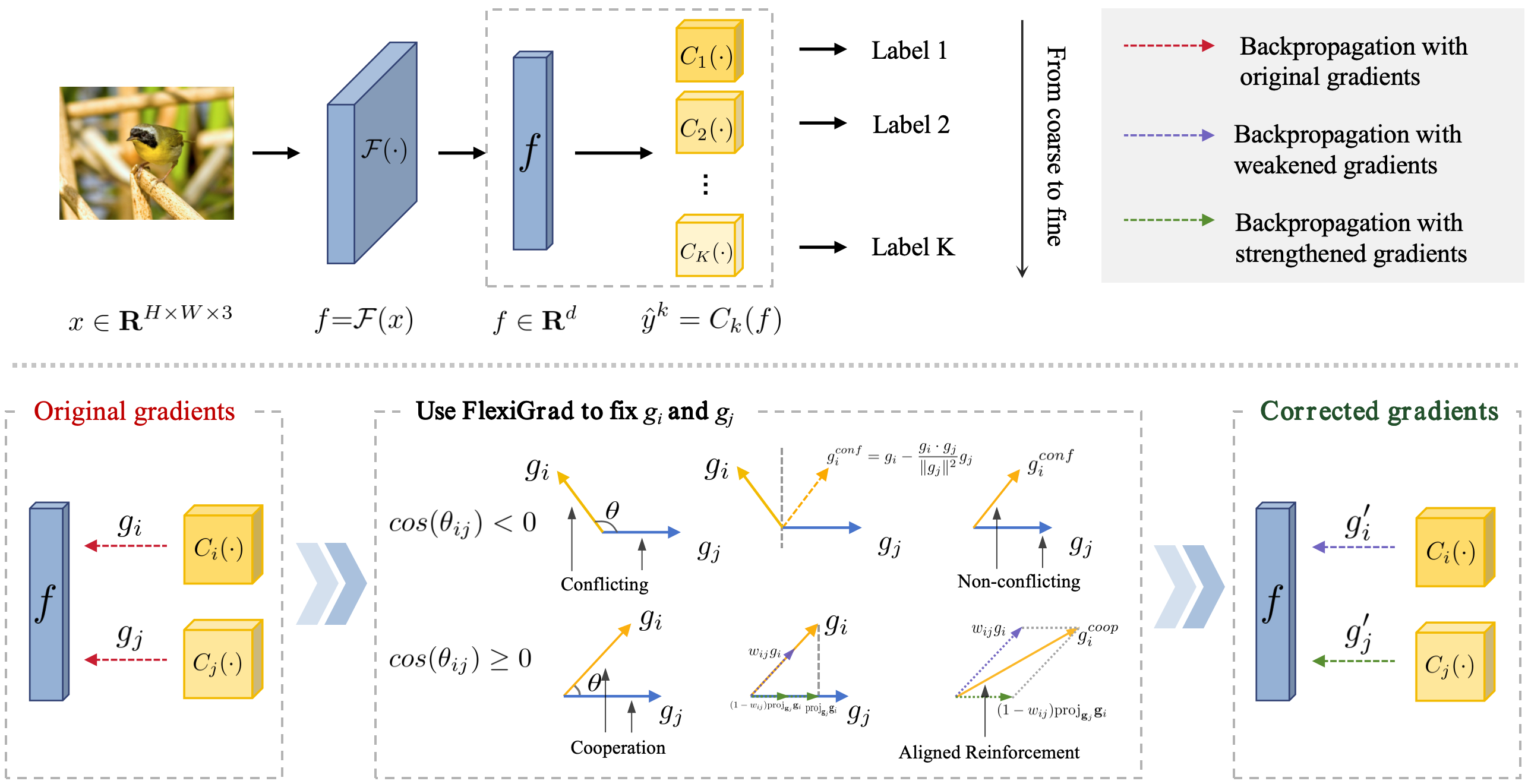}
\caption{Overview of our hierarchy classification model and FlexiGrad method. \(g_i\) and \(g_j\) represent the gradients backpropagating from classifier \(C_i\) and classifier \(C_j\) ($i,j\in K$), which perform classification tasks of coarser granularity level and finer granularity level, respectively. \(\theta\) represents the angle between \(g_i\) and \(g_j\).} \label{fig:2}
\vspace{-3mm}
\end{figure}
\section{Method}
\label{sec:method}

Hierarchical fine-grained classification requires predicting labels at several semantic levels.  
These levels follow a hierarchical structure rather than a set of independent objectives.  
Coarse classifiers encourage global structure and fine classifiers emphasise subtle discriminative cues.  
When jointly optimised, their gradients often misalign, producing hierarchical gradient conflict that disturbs the descent path of the shared backbone.  
FlexiGrad resolves this issue by constructing a hierarchy-aware gradient update that suppresses harmful interactions and strengthens beneficial ones, enabling a coherent coarse-to-fine optimisation process (Figure~\ref{fig:2}).
\vspace{-2mm}

\subsection{Hierarchical Optimisation Objective}

Given an image \(x\) with labels \(\{y^{1}, y^{2}, \dots, y^{K}\}\), the backbone \(\mathcal{F}\) extracts a feature \(f\), and each classifier \(C_k\) predicts \(\hat y^{k}\).  
The standard objective minimises the sum of cross-entropy losses  
\begin{equation}
\mathcal{L}_{\text{total}} = \sum_{k=1}^{K} \mathcal{L}_k.
\end{equation}
Its gradient is the raw sum  
\begin{equation}
g_{\text{raw}} = \sum_{k=1}^{K} g_k,
\end{equation}
where \(g_k = \nabla_\theta \mathcal{L}_k\).  
This aggregation ignores the semantic dependencies among granularity levels and treats all supervision as equally compatible.  
In practice, coarse and fine tasks often push the parameters toward inconsistent regions, so the naive update direction becomes suboptimal.

FlexiGrad retains the same optimisation target \(\mathcal{L}_{\text{total}}\) but reshapes the descent direction. Algorithm~\ref{alg:flexigrad} provides a comprehensive breakdown of the FlexiGrad execution flow within a single training iteration.
Instead of using \(g_{\text{raw}}\), FlexiGrad computes a hierarchy-consistent update  
\begin{equation}
g_{\text{FG}} = \sum_{k=1}^{K} g_k^{\text{FG}},
\end{equation}
where each corrected gradient \(g_k^{\text{FG}}\) incorporates conflict resolution or cooperative reinforcement based on its relationship with other tasks.

\subsection{Gradient Interaction Modelling}
For two granularity levels \(i\) and \(j\), we measure their interaction using cosine similarity  
\begin{equation}
\cos(\theta_{ij}) = \frac{g_i \cdot g_j}{\|g_i\| \|g_j\|}.
\end{equation}
Negative similarity indicates a conflict, and positive similarity indicates partial agreement.  
FlexiGrad uses this value not as a binary signal but as a continuous measure of consistency.  
The corrected gradient \(g_i^{\text{FG}}\) is obtained by iteratively interacting with all other tasks.

\begin{figure}[t] 
    \vspace{-4mm}
    \centering
    \begin{minipage}{0.75\textwidth} 
    \vspace{-4mm}
        \begin{algorithm}[H] 
            \caption{FlexiGrad Update Rule}
            \label{alg:flexigrad}
            \textbf{Require:} Model parameters $\theta$, task minibatch $\mathcal{B} = \{\mathcal{T}_k\}$
            \begin{spacing}{1.3}
            \begin{algorithmic}[1]
                \STATE $g_k \leftarrow \nabla_\theta \mathcal{L}_k(\theta) \quad \forall k$
                \STATE $g_k^{\text{FG}} \leftarrow g_k \quad \forall k$
                \FOR{$\mathcal{T}_i\in\mathcal{B}$}
                    \FOR{$\mathcal{T}_j \in \{\mathcal{T}_k \mid k > i\}$ in sequential order}
                        \STATE Compute $\cos(\theta_{ij}) = \frac{\langle g_i, g_j \rangle}{\|g_i\| \|g_j\|}$
                        \IF{ $\cos(\theta_{ij}) < 0$ } 
                            \STATE // \textit{conflicting}
                            \STATE $g_i^{\text{FG}} \leftarrow g_i^{\text{FG}}-\frac{\langle g_i, g_j \rangle}{\|g_j\|^2+\epsilon} g_j$  \hfill 
                        \ELSE
                            \STATE // \textit{cooperative}
                            \STATE $w_{ij}=\frac{1+\cos(\theta_{ij})}{2}$  \hfill 
                            \STATE $g_i^{\text{FG}} \leftarrow w_{ij}g_i^{\text{FG}}+(1-w_{ij})\frac{\langle g_i, g_j \rangle}{\|g_j\|^2+\epsilon} g_j$  \hfill 
                        \ENDIF
                    \ENDFOR
                \ENDFOR
                \STATE \textbf{return} Update $\Delta\theta=g_{final}^{\text{FG}}=\sum_ig_i^{\text{FG}}$
            \end{algorithmic}
            \end{spacing}
        \end{algorithm}
    \end{minipage}
\vspace{-3mm}
\end{figure}

\subsection{Conflict Resolution}

When \(\cos(\theta_{ij}) < 0\), gradients from the two levels pull the backbone in inconsistent directions.  
Removing the entire influence of one task would discard useful information, so FlexiGrad removes only the harmful component  
\begin{equation}
g_i^{\text{conf}} = g_i - \frac{g_i \cdot g_j}{\|g_j\|^2 + \epsilon} g_j.
\end{equation}
This preserves the informative part of \(g_i\) while preventing destructive interference.  
The update remains close to the original objective but avoids regions conflicting with the hierarchy.

\subsection{Cooperation Reinforcement}

Granularity levels often share partially aligned supervision.  
Instead of treating this as perfect agreement, FlexiGrad strengthens it proportionally using a smooth alignment weight  
\begin{equation}
w_{ij} = \frac{1 + \cos(\theta_{ij})}{2}.
\end{equation}
The reinforced gradient becomes  
\begin{equation}
g_i^{\text{coop}}
= w_{ij} g_i + (1 - w_{ij}) \frac{g_i \cdot g_j}{\|g_j\|^2 + \epsilon} g_j.
\end{equation}
This encourages the model to move in directions supported consistently by multiple levels while still preserving task-specific fine-grained cues.  
The continuous weighting naturally reflects the hierarchical dependency: coarse supervision stabilises global representation, and fine supervision guides detailed discrimination.

\subsection{Full Gradient Modulation}

For each task \(i\), FlexiGrad starts with its raw gradient \(g_i\) and sequentially applies the conflict or cooperation rule with all other tasks \(j \ne i\).  
After this process, we obtain corrected gradients \(g_i^{\text{FG}}\), and the final update becomes  
\begin{equation}
g_{\text{FG}} = \sum_{i=1}^{K} g_i^{\text{FG}}.
\end{equation}
FlexiGrad introduces no parameters, requires no modifications to the architecture and integrates seamlessly into any optimiser by replacing the raw gradient with \(g_{\text{FG}}\).

\subsection{Theoretical Interpretation}

The corrected update direction \(g_{\text{FG}}\) remains a valid descent direction for the total objective.  
When gradients conflict, removing the opposing component yields  
\begin{equation}
g_i^{\text{conf}} \cdot g_i = \|g_i\|^2 - \frac{(g_i \cdot g_j)^2}{\|g_j\|^2} \geq 0,
\end{equation}
so the update continues to decrease \(\mathcal{L}_i\).  
When gradients are partially aligned, reinforcement produces  
\begin{equation}
g_i^{\text{coop}} \cdot g_i = w_{ij} \|g_i\|^2 + (1 - w_{ij}) \frac{(g_i \cdot g_j)^2}{\|g_j\|^2} > 0,
\end{equation}
so the descent direction reduces \(\mathcal{L}_i\) while benefiting from agreement with \(\mathcal{L}_j\).  
Since each corrected gradient remains a non-ascending direction for its own task, their sum \(g_{\text{FG}}\) constitutes a principled surrogate descent for \(\mathcal{L}_{\text{total}}\).  
FlexiGrad therefore retains the original optimisation objective but yields a smoother, hierarchy-consistent trajectory in parameter space.

\subsection{Summary}

FlexiGrad introduces a continuous, hierarchy-aware gradient modulation mechanism that captures both conflict and cooperation across granularity levels.  
It preserves the discriminative power of fine-level supervision while stabilizing global representation through coarse supervision, and reshapes the descent path to align with the semantic structure of hierarchical classification.

\section{Experiment}
\label{sec:experiments}

\subsection{Experimental Setup}
\textbf{Datasets and Evaluation:} We evaluate FlexiGrad on three hierarchical FGVC benchmarks: \textbf{CUB-200-2011}~\cite{wah2011caltech}: 11,877 bird images annotated at order, family and species levels.  
Its deep taxonomy and subtle species differences expose strong coarse-to-fine gradient interference.
\textbf{FGVC-Aircraft}~\cite{maji2013fine}: 10,000 images from 30 makers, 70 families and 100 variants.  
Fine-level distinctions are extremely subtle, making it a natural stress test for gradient alignment.
\textbf{Stanford Cars}~\cite{krause2013car}: 8,144 images covering 196 models.  
Rigid shapes but large colour variations create inconsistent coarse-level cues.
We report accuracy at each granularity and the average, averaged over three runs.

\textbf{Implementation:} FlexiGrad is model-agnostic.  
We use ResNet50, Swin-B and ViT-B/16 pretrained on ImageNet~\cite{deng2009imagenet}.  
Images are resized to \(224\times224\); standard augmentations are used.  
Training uses SGD for 100 epochs with weight decay \(5\times10^{-4}\).  
All models follow identical optimisation settings to ensure comparability.

\textbf{Baselines:} We compare against representative strategies for hierarchical multi-task learning: 
\textbf{Vanilla\_single}: shared backbone without conflict handling, \textbf{Vanilla\_multi}: separate backbones, removing all gradient interaction, \textbf{PCGrad}~\cite{yu2020gradient}: projection-based conflict removal, \textbf{FGoN}~\cite{chang2021your}: architecture-level gradient blocking, and \textbf{Ours}: FlexiGrad applied to Vanilla\_single. To test generality, we also apply FlexiGrad as a plug-in to four strong FGVC models (NTS~\cite{yang2018NTS}, PMG~\cite{du2020pmg}, MPSA~\cite{wang2024mpsa} and ACC-ViT~\cite{zhang2024accvit}), producing \textbf{Ours\_NTS}, \textbf{Ours\_PMG}, \textbf{Ours\_MPSA} and \textbf{Ours\_ACC-ViT}.
\vspace{-1mm}

\begin{table}[t]
 \caption{Performance comparison on three datasets. 
 \vspace{1mm}
 \textbf{\textcolor{blue}{Blue}} indicates the best results among baseline methods, while \textbf{\textcolor{red}{Red}} highlights the best performance among SOTA approaches and our plug-in versions.} 
 \vspace{4mm}
 \label{tab:1} 
 \resizebox{1\linewidth}{!}{ 
 \renewcommand{\arraystretch}{1.72}
 \begin{tabular}{l|cccc|cccc|ccc} 
 \toprule
 \multirow{2}{*}{\textbf{Method}} & 
 \multicolumn{4}{c|}{\textbf{CUB-200-2011}} & 
 \multicolumn{4}{c|}{\textbf{FGVC-Aircraft}} & 
 \multicolumn{3}{c}{\textbf{Stanford Cars}} \\ 
 \cmidrule{2-12} 
  & \textbf{order\_acc} & \textbf{family\_acc} & \textbf{species\_acc} & \textbf{avg\_acc} & 
   \textbf{maker\_acc} & \textbf{family\_acc} & \textbf{model\_acc} & \textbf{avg\_acc} & 
   \textbf{maker\_acc} & \textbf{model\_acc} & \textbf{avg\_acc}\\ 
 \midrule
 \midrule
 \textbf{Vanilla\_single} 
 & 94.97 ± 0.27 & 87.74 ± 0.11 & 73.82 ± 0.31 & 85.51 
 & 94.59 ± 0.03 & 92.13 ± 0.42 & 86.93 ± 0.21 & 91.22 
 & 95.03 ± 0.25 & 88.59 ± 0.13 & 91.81 \\ 
 \textbf{Vanilla\_multi~\cite{chang2021your}} 
 & 95.13 ± 0.53 & 89.70 ± 0.13 & 78.31 ± 0.35 & 87.71 
 & 90.69 ± 0.48 & 89.23 ± 0.23 & 88.10 ± 0.10 & 89.34 
 & 95.24 ± 0.20 & 89.14 ± 0.16 & 92.19\\ 
 \textbf{PCGrad} 
 & 96.70 ± 0.14 & 90.14 ± 0.32 & 78.07 ± 0.19 & 88.30 
 & 94.30 ± 0.28 & 92.67 ± 0.05 & 87.80 ± 0.36 & 91.59 
 & 95.19 ± 0.16 & 89.15 ± 0.27 & 92.17 \\ 
 \textbf{FGoN~\cite{chang2021your}} 
 & 96.37 ± 0.16 & 90.39 ± 0.15 & 77.95 ± 0.04 & 88.24 
 & 93.04 ± 0.25 & 90.73 ± 0.19 & 88.35 ± 0.18 & 90.71 
 & 95.58 ± 0.06 & 89.66 ± 0.16 & 92.62\\ 
 \textbf{Ours} 
 & \textbf{\textcolor{blue}{96.94 ± 0.10}} & \textbf{\textcolor{blue}{90.85 ± 0.33}} & \textbf{\textcolor{blue}{79.79 ± 0.41}} & \textbf{\textcolor{blue}{89.19}} 
 & \textbf{\textcolor{blue}{94.56 ± 0.06}} & \textbf{\textcolor{blue}{93.00 ± 0.21}} & \textbf{\textcolor{blue}{88.76 ± 0.15}} & \textbf{\textcolor{blue}{92.11}} 
 & \textbf{\textcolor{blue}{95.80 ± 0.34}} & \textbf{\textcolor{blue}{90.58 ± 0.18}} & \textbf{\textcolor{blue}{93.19}} \\ 
 \toprule 
 \textbf{NTS} 
 & 97.60 ± 0.15 & 93.55 ± 0.18 & 83.75 ± 0.20 & 91.63 
 & 95.41 ± 0.25 & 93.85 ± 0.12 & 88.60 ± 0.14 & 92.62 
 & 95.20 ± 0.21 & 91.98 ± 0.19 & 93.59 \\ 
 \textbf{Ours\_NTS} 
 & 97.91 ± 0.11 & 93.98 ± 0.22 & 84.01 ± 0.15 & 91.97 
 & 95.68 ± 0.19 & 94.16 ± 0.18 & 88.95 ± 0.11 & 92.93 
 & 95.62 ± 0.24 & 92.61 ± 0.15 & 94.12 \\ 
 \hline 
 \textbf{PMG} 
 & 97.93 ± 0.09 & 93.35 ± 0.42 & 83.32 ± 0.17 & 91.53 
 & 95.88 ± 0.30 & 93.79 ± 0.04 & 87.85 ± 0.29 & 92.51 
 & 94.93 ± 0.06 & 91.29 ± 0.45 & 93.11 \\ 
 \textbf{Ours\_PMG} 
 & 98.21 ± 0.56 & 94.21 ± 0.25 & 84.36 ± 0.13 & 92.26 
 & 96.22 ± 0.08 & 94.06 ± 0.61 & 89.25 ± 0.10 & 93.18 
 & \textbf{\textcolor{red}{96.25 ± 0.33}} & \textbf{\textcolor{red}{93.18 ± 0.12}} & \textbf{\textcolor{red}{94.72}} \\ 
 \hline 
 \textbf{ACC-ViT} 
 & 99.50 ± 0.08 & 97.10 ± 0.15 & 89.18 ± 0.11 & 95.26 
 & 96.80 ± 0.19 & 94.75 ± 0.14 & 86.55 ± 0.22 & 92.70 
 & 95.45 ± 0.15 & 87.10 ± 0.25 & 91.28 \\ 
 \textbf{Ours\_ACC-ViT}
 & \textbf{\textcolor{red}{99.77 ± 0.05}} & 97.52 ± 0.12 & \textbf{\textcolor{red}{89.45 ± 0.14}} & 95.58 
 & 97.01 ± 0.15 & 95.02 ± 0.10 & 86.91 ± 0.18 & 92.98 
 & 95.67 ± 0.18 & 87.36 ± 0.21 & 91.52 \\ 
 \hline 
 \textbf{MPSA} 
 & 99.57 ± 0.09 & 97.90 ± 0.17 & 88.82 ± 0.04 & 95.43 
 & 97.03 ± 0.28 & 94.93 ± 0.11 & 89.26 ± 0.20 & 93.74 
 & 95.82 ± 0.13 & 91.05 ± 0.31 & 93.44 \\ 
 \textbf{Ours\_MPSA} 
 & 99.71 ± 0.06 & \textbf{\textcolor{red}{98.10 ± 0.12}} & 89.37 ± 0.16 & \textbf{\textcolor{red}{95.73}} 
 & \textbf{\textcolor{red}{97.19 ± 0.27}} & \textbf{\textcolor{red}{95.07 ± 0.21}} & \textbf{\textcolor{red}{89.77 ± 0.15}} & \textbf{\textcolor{red}{94.01}} 
 & 96.01 ± 0.20 & 91.47 ± 0.39 & 93.74 \\ 
 \bottomrule
 \end{tabular} 
 } 
 \vspace{-5mm}
 \end{table}

\subsection{Comparisons with State-of-the-Art Methods}

Table~\ref{tab:1} compares against state-of-the-art methods.  
Across all datasets, FlexiGrad consistently improves accuracy at every granularity and across all backbones.  
More importantly, the pattern of improvement directly reflects the optimisation behaviour predicted by our hierarchy-aware formulation.

First, the largest gains emerge at the finest granularity.  
Species-and model-level classifiers benefit substantially more than coarse levels, indicating that FlexiGrad effectively suppresses the strong coarse-to-fine gradient interference that typically harms fine-grained recognition.   
This matches our core hypothesis: finer tasks suffer the most misalignment and therefore gain the most from continuous conflict removal and cooperation reinforcement.

Second, this improvement persists across model families.  
FlexiGrad enhances both CNN-based FGVC models (NTS, PMG) and transformer-based architectures (ACC-ViT, MPSA), demonstrating that hierarchy-aware gradient modulation acts on the optimisation dynamics rather than architectural priors.  
The consistent gains also indicate that task-level interference cannot be fully resolved by stronger feature extractors or attention modules alone.  
Even models with explicit attention mechanisms still exhibit measurable gains, suggesting that task-level interference remains a limiting factor that architecture alone does not resolve.

Third, FlexiGrad significantly outperforms projection-based (PCGrad) and gradient-blocking (FGoN) methods by a clear margin.  
Pure projection often removes useful components, while strict blocking prevents beneficial top-down and bottom-up interactions.  
In contrast, FlexiGrad distinguishes conflicting and cooperative cases through a continuous alignment measure, which matches the mixed nature of hierarchical supervision.  
FlexiGrad’s continuous alignment strategy preserves helpful cooperation while eliminating only the harmful part, leading to a consistently stronger optimisation trajectory.

Together, these results show that FlexiGrad delivers robust, architecture-agnostic improvements by reshaping the gradient landscape to better reflect the hierarchical structure of multi-granularity FGVC.


\section{Further Analysis}
\label{sec:analysis}

\subsection{Visualization}

\begin{figure}[t]
\includegraphics[width=\textwidth]{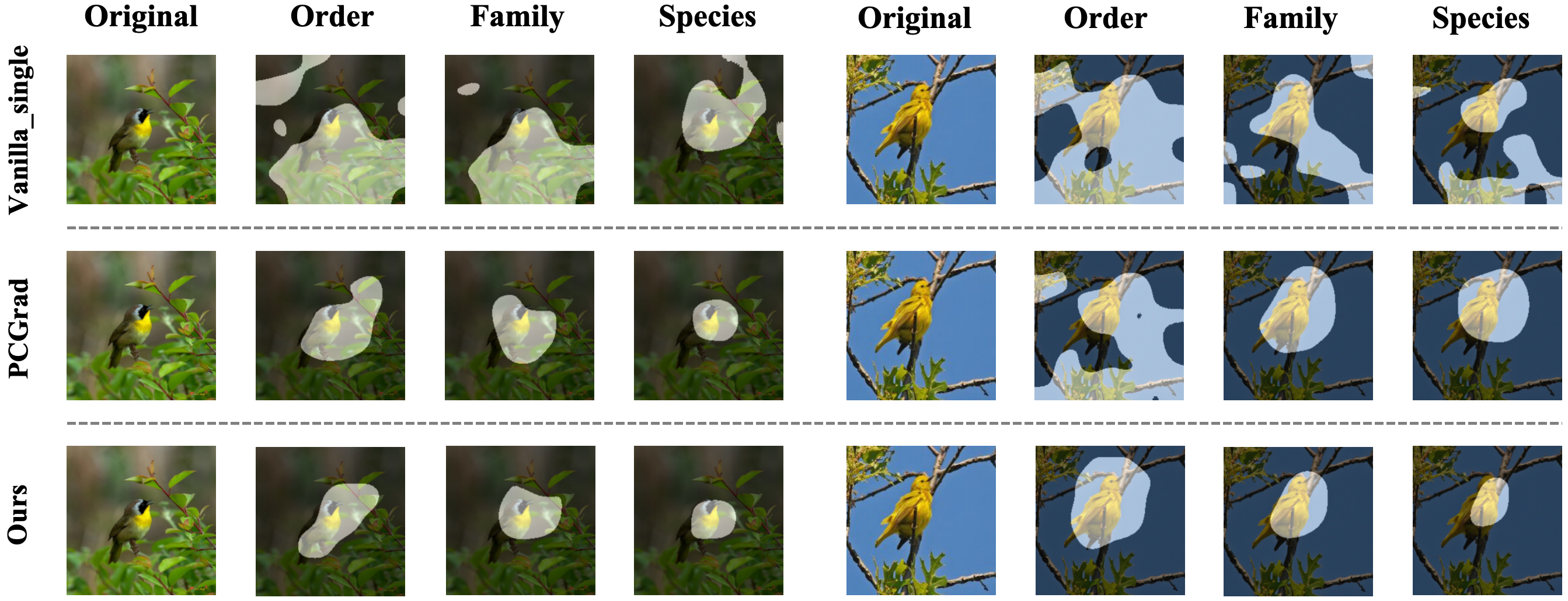}
\caption{Visualizations of Vanilla\_single with PCGrad and Ours at Order, Family, and Species granularities on CUB-200-2011.} \label{fig:3}
\vspace{-4mm}
\end{figure}

We visualise how FlexiGrad reshapes hierarchical optimisation through three complementary analyses.  
Across all visualisations, a consistent pattern emerges: FlexiGrad produces coarse-to-fine feature refinement that matches the structure of the hierarchy.

\noindent \textbf{Hierarchical attention progression.}  
Figure~\ref{fig:3} shows Grad-CAM~\cite{selvaraju2017grad} maps at order, family and species levels.  
Vanilla and PCGrad often produce diffuse activations or misaligned focus across levels, reflecting inconsistent gradient signals.  
FlexiGrad generates a clear coarse-to-fine progression: broad body regions at the order level, more semantically coherent structures at the family level, and sharply localised fine details at the species level.  
This behaviour indicates stabilised coarse gradients and strengthened fine-level cues, consistent with our gradient modulation design.

\begin{figure}[t]
\includegraphics[width=\textwidth]{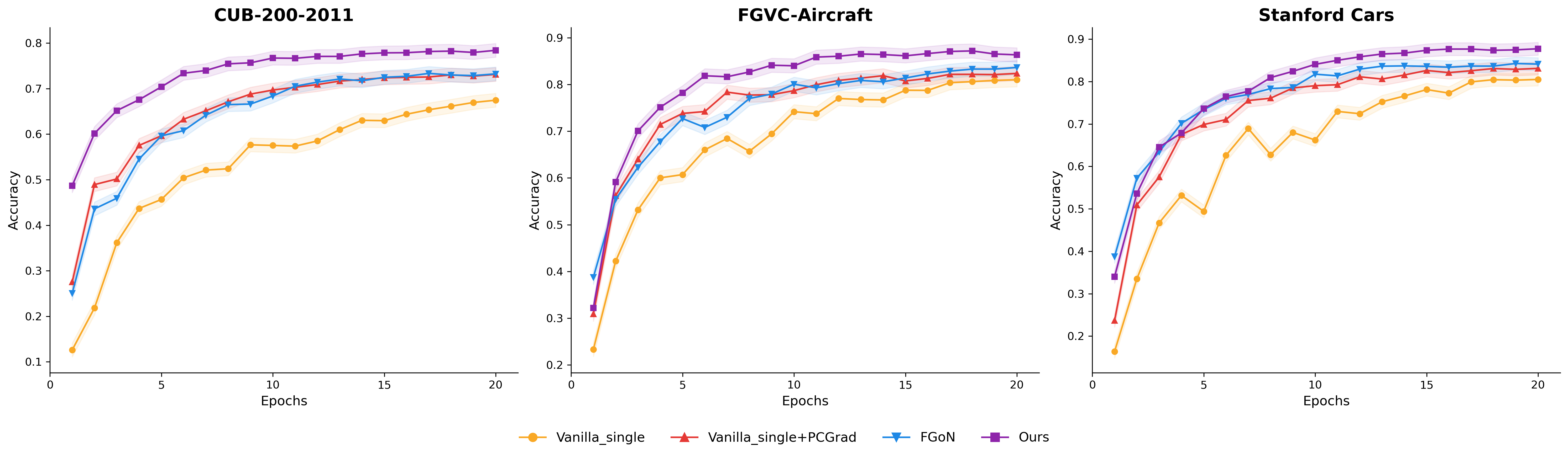}
\caption{Accuracy performance comparison of Vanilla\_single, Vanilla\_single+PCGrad, FGoN, and our method on CUB-200-2011 during the first 20 epochs.} \label{fig:5}
\vspace{-1mm}
\end{figure}

\noindent \textbf{Optimisation dynamics.}  
Figure~\ref{fig:5} compares the first 20 training epochs.  
FlexiGrad converges faster and exhibits significantly reduced oscillation.  
This suggests a smoother, more coherent descent direction, whereas Vanilla and PCGrad suffer from gradient tug-of-war between levels.  
The separation between different methods may become more pronounced in scenarios with deeper hierarchies and more intense gradient conflicts.

\begin{figure}[t]
\includegraphics[width=\textwidth]{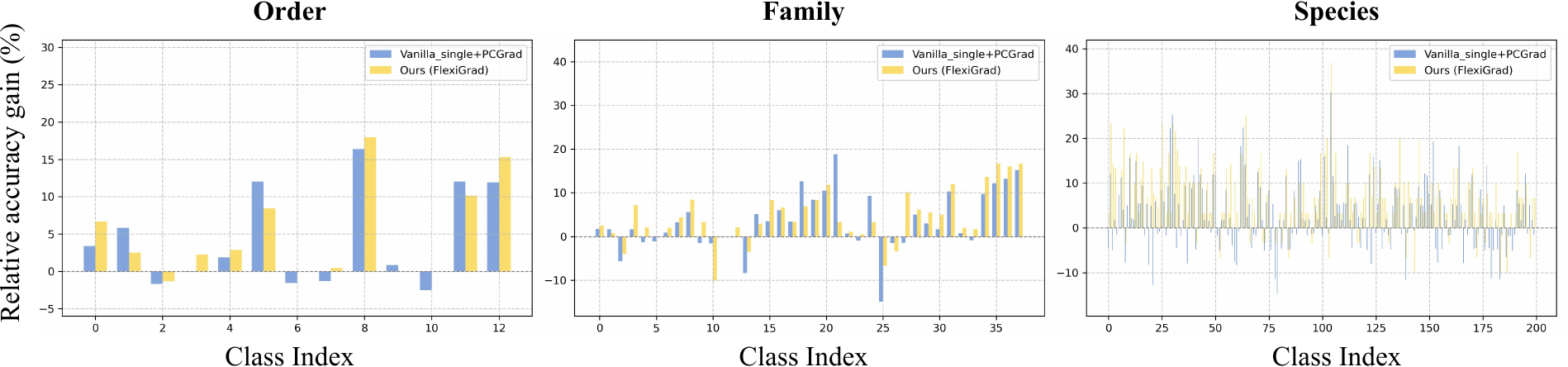}
\caption{Relative accuracy gain versus Vanilla\_single between Vanilla\_single with PCGrad and our method within Order, Family and Species hierarchy on the CUB-200-2011 dataset.} \label{fig:4}
\vspace{-2mm}
\end{figure}

\noindent \textbf{Per-class improvement structure.}  
Figure~\ref{fig:4} highlights class-wise gains relative to PCGrad.  
FlexiGrad improves the majority of classes and the largest gains occur in high-confusion species, precisely where coarse-to-fine interference is typically the strongest.  
The improvement distribution aligns with our intuition: FlexiGrad amplifies useful inter-level agreement and suppresses destructive interference, giving the biggest advantage in the hardest regions of the taxonomy.

Together, these visual analyses show that FlexiGrad not only improves accuracy but also induces a coherent hierarchical optimisation behaviour that is visible at the feature, training and category levels.

\subsection{Ablation of Core Components}

Table~\ref{tab:tabVI} evaluates the two components separately.  
Using only the conflict resolution mechanism increases the average accuracy to 88.30 percent, which demonstrates that removing the opposing gradient component effectively reduces negative coarse-to-fine interference.  
Using only the weighting mechanism increases the accuracy to 88.86 percent by amplifying gradients that move in partially consistent directions.  
Combining the two components gives the best result of 89.19 percent.  
The improvement shows that eliminating harmful interference and reinforcing beneficial alignment are both necessary and that their effects complement one another. 

\vspace{-1mm}

\subsection{Design of the Weighting Strategy}
To verify the effectiveness of our linear mapping, we evaluate four alternative strategies (Table~\ref{tab:wi}). The \textbf{Temperature-scaled Sigmoid ($w_1$)} fails due to saturation and "hard" switching that misses subtle inter-task agreements. The \textbf{Exponential Mapping ($w_2$)} is overly sensitive to gradient variations, leading to optimization instability. For structural designs, \textbf{Granularity-aware Scaling ($w_3$)} imposes a rigid, non-adaptive prior, while \textbf{Learnable Parameters ($w_4$)} introduce unstable bi-level optimization problem. 

Experimental results demonstrate that our proposed $w_{ij} = (1+\cos\theta_{ij})/2$ consistently achieves the highest accuracy across all granularities. By providing a parameter-free link between alignment and update strength, this linear mapping ensures stable reinforcement proportional to task consistency.

\begin{table}[t]
\centering
\vspace{2mm}
\caption{Ablation study of the proposed components on the CUB-200-2011 dataset. ``\checkmark'' indicates the corresponding module is enabled.}
\vspace{2mm}
\setlength{\tabcolsep}{5pt}
\renewcommand{\arraystretch}{1.2}
\fontsize{6pt}{7pt}\selectfont
\resizebox{0.74\linewidth}{!}{
\begin{tabular}{c c| c c c c}
\toprule
\textbf{Projection} & \textbf{Reweight} & \textbf{order\_acc} & \textbf{family\_acc} & \textbf{species\_acc} & \textbf{avg\_acc} \\
\midrule
&  &  94.97 & 87.74 & 73.82 & 85.51 \\
\checkmark &  & 96.70 & 90.14 & 78.07 & 88.30 \\
& \checkmark & 96.82 & 90.54 & 79.21 & 88.86 \\
\checkmark & \checkmark & \textbf{96.94} & \textbf{90.85} & \textbf{79.79} & \textbf{89.19} \\
\bottomrule
\end{tabular}
}
\label{tab:tabVI}
\vspace{-1mm}
\end{table}

\begin{table}[t]
\centering
\vspace{2mm}
\caption{Ablation study on different $w_i$ design strategies on the CUB-200-2011 dataset. ($w_1$: Temperature-scaled Sigmoid, 
$w_2$: Exponential Mapping, 
$w_3$: Granularity-aware Scaling, 
$w_4$: Learnable Parameters.)}
\label{tab:wi}
\vspace{2mm}
\setlength{\tabcolsep}{5pt}
\renewcommand{\arraystretch}{1.2}
\fontsize{6pt}{7pt}\selectfont
\resizebox{0.68\linewidth}{!}{
\begin{tabular}{c|cccc}
\toprule
\textbf{Design Strategy} & \textbf{order\_acc} & \textbf{family\_acc} & \textbf{specie\_acc} & \textbf{avg\_acc} \\
\midrule
baseline  & ~\textbf{96.94} & ~\textbf{90.85} & ~\textbf{79.79} & ~\textbf{89.19} \\
$w_1$  &  96.60 & 90.08 & 79.31 & 88.66 \\
$w_2$  & 96.63 & 90.54 & 79.41 & 88.86 \\
$w_3$  & 96.75 & 90.65 & 79.38 & 88.93 \\
$w_4$  & 96.58 & 90.72 & 79.53 & 88.94 \\
\bottomrule
\end{tabular}
}
\vspace{-4mm}
\end{table}

\subsection{Efficiency and Computational Overhead}
FlexiGrad introduces only a small computational overhead.  
The modulation is performed on task-specific gradients before the final optimizer update, and therefore mainly relies on lightweight vector operations without extra forward passes or additional network parameters.  
As shown in Table~\ref{tab:time}, FlexiGrad achieves the highest average accuracy on CUB-200-2011, while its training time remains close to existing methods.

\begin{table}[t]
\centering
\caption{The computational overhead and average accuracy comparisons on CUB-200-2011.}
\label{tab:time}
\vspace{1mm}
\setlength{\tabcolsep}{5pt}
\renewcommand{\arraystretch}{1.3}
\fontsize{6pt}{7pt}\selectfont
\resizebox{0.8\linewidth}{!}{
\begin{tabular}{c|ccc}
\toprule
\textbf{Method} & \textbf{Time/Run (min)} & \textbf{Overhead vs. Vanilla} & \textbf{avg\_acc} \\
\midrule
Vanilla\_single  & 41.30 & - & 85.51\\
FGoN  &  42.11 & +2.0\% & 88.24\\
Vanilla\_single+PCGrad  & 43.98 & +6.5\% & 88.30\\
Ours  & 44.37 & +7.4\% & 89.19\\
\bottomrule
\end{tabular}
}
\vspace{-2mm}
\end{table}

The results demonstrate that the minor increase in processing time relative to PCGrad is well-justified. By introducing a sequential weighting function $w_{ij}$ to manage hierarchical conflicts, FlexiGrad yields a significant +0.89\% absolute gain in average accuracy over the strongest baseline (PCGrad) with only a +0.9\% increase in time. This marginal trade-off confirms that FlexiGrad is both effective and computationally practical for real-world fine-grained recognition tasks.

\begin{figure}
\centering
\includegraphics[width=0.78\textwidth]{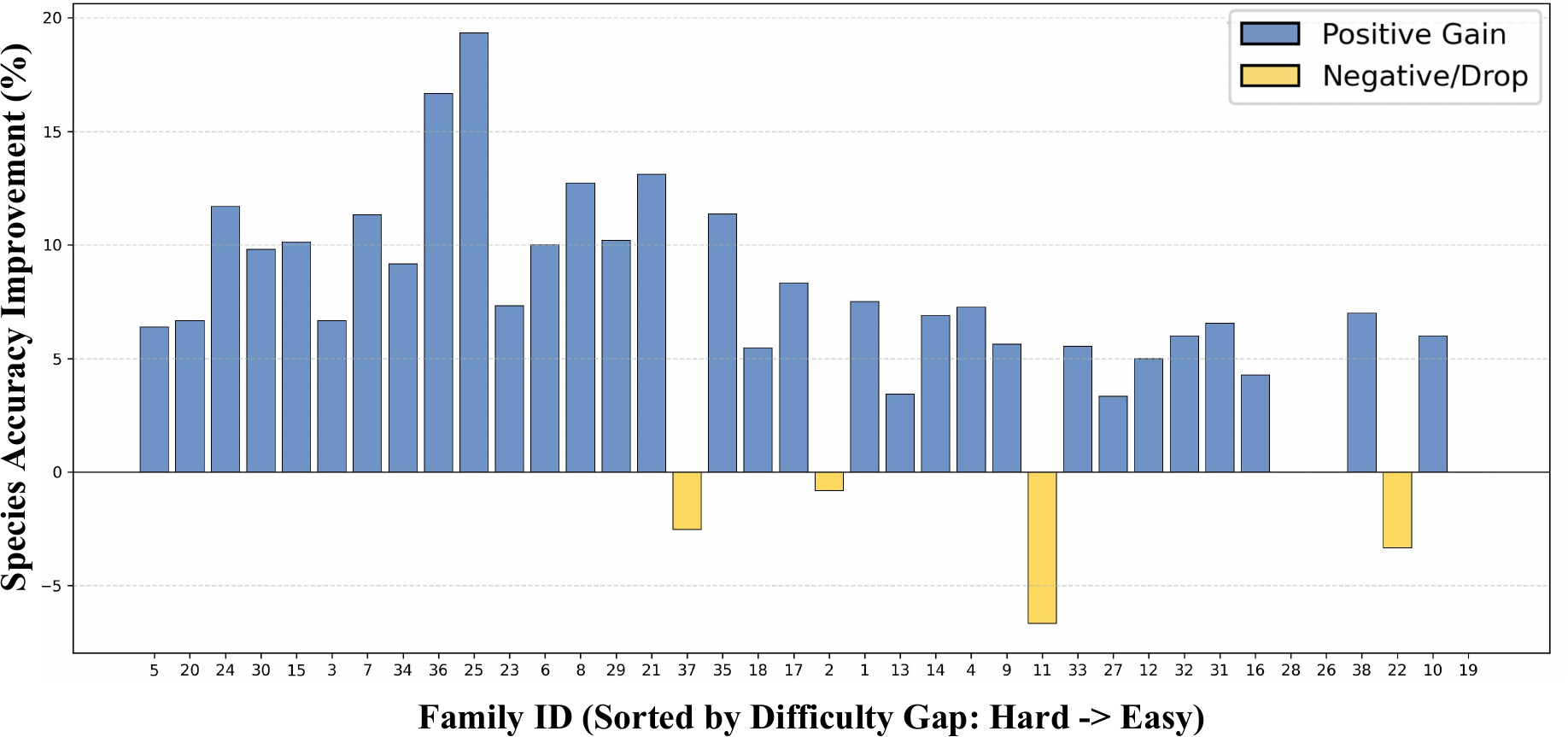}
\caption{Species-level accuracy improvement(\%) versus Vanilla\_single on the CUB-200-2011 dataset across different bird families.} \label{fig:6}
\vspace{-5mm}
\end{figure}

\subsection{Performance on Challenging Fine-Grained Subtrees}

To assess FlexiGrad under the most demanding conditions, we conduct a hard-scenario evaluation focusing on specific taxonomic subtrees. We group samples by family and rank these families based on the discrepancy between family-level and species-level accuracies in the baseline model. A large gap indicates that while the model captures coarse global patterns, it fails to resolve the subtle, discriminative features required for species-level distinction due to overwhelming hierarchical interference.

As illustrated in Figure~\ref{fig:6}, FlexiGrad yields substantial accuracy gains across the majority of families, with improvements most pronounced in the high-difficulty regions of the taxonomy. Notably, we observe a maximum species-level improvement of 19.33\% in families where the baseline's performance was bottlenecked. By suppressing destructive interactions and reinforcing aligned cross-level signals, FlexiGrad enables the backbone to learn a coherent representation even within these challenging subtrees where standard joint optimization fails to converge to an optimal local minimum.

\section{Conclusion}

This work addressed the core optimisation problem in hierarchical fine-grained classification, where coarse and fine levels generate misaligned gradients that disrupt learning.  
FlexiGrad provides a simple solution by removing the conflicting component of each gradient while reinforcing the consistent part, resulting in a coherent hierarchy-aware update direction.  
The method preserves the benefits of cross-level supervision, suppresses the harmful interference that typically harms fine-grained recognition and produces coarse-to-fine behaviour that matches the structure described in our introduction.  
Experiments, visualisations and hard-scenario analyses collectively confirm that FlexiGrad improves performance exactly where hierarchical conflict is strongest.


%
%
%
%




\bibliographystyle{splncs04}
\bibliography{prcv2026references}

\end{document}